\pdfoutput=1

\documentclass[11pt]{article}

\usepackage[final]{acl}

\usepackage{times}
\usepackage{latexsym}
\usepackage{graphicx}
\usepackage{amsmath} 
\usepackage{enumitem}
\usepackage{amsfonts}
\usepackage{booktabs}

\usepackage[T1]{fontenc}

\usepackage[utf8]{inputenc}

\usepackage{microtype}

\usepackage{inconsolata}

\title{Lacuna Inc. at SemEval-2026 Task 4: Structurally Gated State-Space Models for Disentangling Narrative Similarity}

\usepackage{inconsolata}
\usepackage{hyperref}
\usepackage{fontawesome5}

\author{Aleksey Kudelya\textsuperscript{\faCrow}, 
        Rafif Alshawi\textsuperscript{\faCrow},
        Alexander Shirnin\textsuperscript{\faCrow}\\
    \textsuperscript{\faCrow}HSE University \\
    \small{
    \textbf{Correspondence:} \href{mailto:ashirnin@hse.ru}{\texttt{ashirnin@hse.ru}}
}
}

\begin{document}
\maketitle
\begin{abstract}

In this paper, we present the Invariant-Variant Disentangled State-Space Model (IVD-SSM), our submission to SemEval-2026 Task 4 on Narrative Story Similarity and Narrative Representation Learning. Evaluating narrative similarity is a profound computational challenge that requires models to look past concrete, superficial elements such as specific names, actors, objects, or settings to isolate and compare abstract patterns of causality and plot progression. To model these extended causal chains without the quadratic bottlenecks of standard Transformers, we leverage a hybrid State-Space Model (Jamba-1.5-Mini). Building upon this backbone, we introduce the Structurally Gated Alignment (SGA) head, a novel, differentiable algorithmic architecture. The SGA head operates on two scales: a heavily strided Macro-path maps the coarse structural skeleton of a story, which then acts as a gating mechanism to filter a full-resolution Micro-path, actively suppressing semantic noise and superficial keyword overlaps. Evaluated on both pairwise comparative judgments (Track A) and dense representation learning (Track B), our approach demonstrates that explicitly disentangling structural invariants from lexical variants provides a robust, principled framework for deep narrative understanding.

\end{abstract}

\section{Introduction}

Narrative similarity is a complex cognitive task that requires distinguishing the "deep structure" of a plot from its surface-level realization. As defined by the SemEval 2026 Task 4 organizers, narrative similarity relies on "abstract patterns of causality and progression" while explicitly "disregarding concrete details" such as names, settings, or specific objects. For computational models, this presents a significant challenge: standard Transformer architectures are often biased towards lexical and semantic overlap, making them susceptible to "spurious correlations" - pairs of stories that share genre-specific vocabulary (e.g., "zombies", "spaceships") but diverge fundamentally in their causal chains.

In this work, we present IVD-SSM (Invariant-Variant Disentangled State-Space Model), a novel architecture designed to align directly with this theoretical definition of narrative similarity. The system must be designed to reflect the core dichotomy of the task: the model must isolate the invariant features (the underlying causal chain and structural progression that define the plot) from the variant features (the superficial, interchangeable details such as specific character names, settings, or genre tropes). We identify that the primary failure mode of standard baselines is the conflation of this invariant plot structure with variant semantic noise. To address this, we employ a predominantly frozen Jamba-1.5-Mini backbone with lightweight QLoRA adapters, leveraging its hybrid Mamba-Transformer architecture to capture long-range causal dependencies, and introduce a mechanism to explicitly disentangle these two signals.

\section{Task Description and Related Work}

\subsection{Task description}
The SemEval 2026 Task 4 presents two challenges centered on narrative understanding \citep{hatzel-etal-2026-semeval}. Track A (Comparative Narrative Similarity) is formulated as a pairwise binary classification task, requiring systems to determine which of two candidate stories is structurally closer to an anchor story. Track B (Narrative Representation Learning) tasks systems with producing dense vector embeddings for individual stories such that the cosine distance between representations mathematically reflects human similarity judgments.

\subsection{Related work}
Prior computational approaches to narrative similarity have ranged from shallow lexical and topic-based matching \citep{chaturvedi2018heard, chun2024aistorysimilarity} to deeper structural representations utilizing event chains and character networks \citep{chambers2008unsupervised, lee2020story, lafhel2024graph, hatzel2024story}. However, as the SemEval-2026 Task 4 organizers highlight, evaluating true narrative similarity requires tracking abstract patterns of causality rather than surface-level overlaps \citep{hatzel-etal-2026-semeval}. Modeling these unstructured, long-form narratives directly via Large Language Models (LLMs) has historically been constrained by sequence length limits. Standard Transformer architectures \citep{vaswani2017attention} impose strict input boundaries due to the $O(N^2)$ quadratic memory and computational complexity inherent to their self-attention mechanisms. While sparse-attention adaptations such as Longformer \citep{beltagy2020longformer} and BigBird \citep{zaheer2020big} artificially extend this context window, they remain computationally heavy during inference and rely on attention approximations that risk fragmenting long-range causal dependencies. To overcome these bottlenecks, we turn to State-Space Models (SSMs). Recent advancements in selective SSMs, specifically the Mamba architecture \citep{gu2023mamba}, offer a linear-time $O(N)$ alternative to self-attention by compressing sequence history into a dynamically updated, fixed-size hidden state. Because SSMs natively track state variables across extended sequences without the massive memory overhead of key-value caches, our chosen backbone-the hybrid Mamba-Transformer model, Jamba-1.5-Mini \citep{lieber2024jambahybridtransformermambalanguage}, is uniquely equipped to capture the plot's "course of action", making it both theoretically and computationally ideal for document-level narrative analysis.

\section{System overview}

\begin{figure*}[t]
    \centering
    \includegraphics[width=\textwidth]{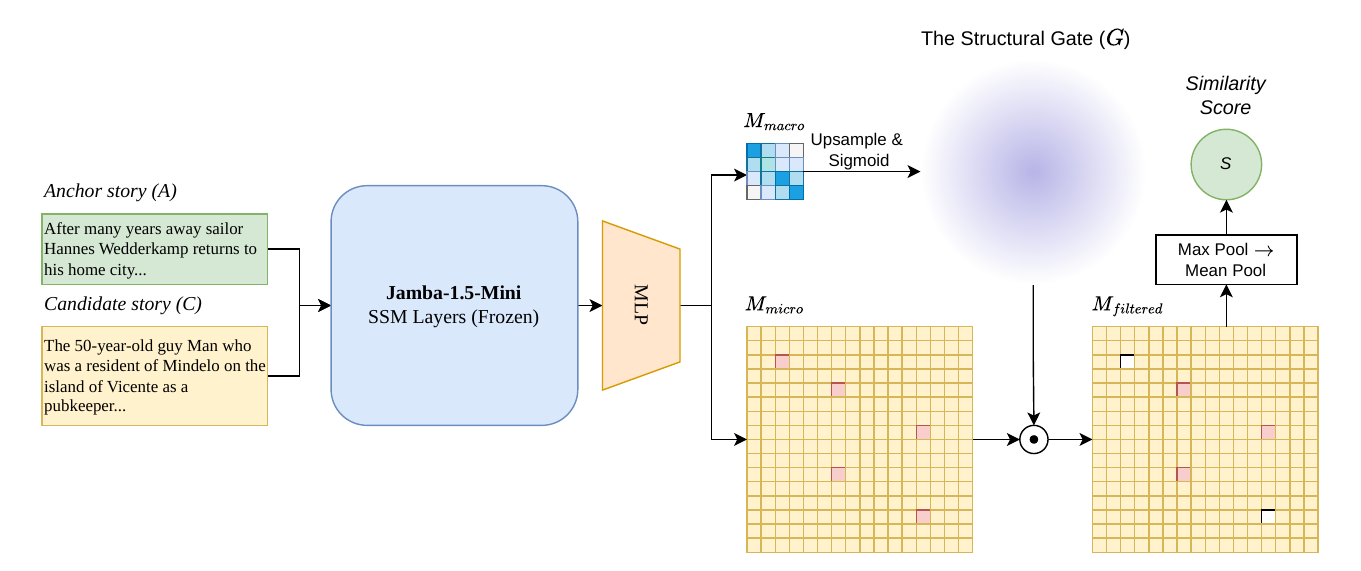}
    \caption{The Invariant-Variant Disentangled State-Space Model (IVD-SSM) architecture. The Jamba-1.5-Mini backbone (predominantly frozen, with lightweight QLoRA adapters) processes candidate and anchor narratives, projecting them into a dual-path alignment head. The heavily strided Macro-path ($M_{macro}$) captures the coarse, invariant plot structure and generates a soft structural mask via upsampling and a Sigmoid activation. This mask acts as a structural gate, applied via element-wise multiplication ($\odot$) to the high-resolution Micro-path ($M_{micro}$). This mechanism explicitly filters out variant semantic noise and adversarial lexical distractors, yielding a filtered final alignment matrix that is pooled into a final similarity score.}
    \label{fig:architecture}
\end{figure*}

Our system, the Invariant-Variant Disentangled State-Space Model (IVD-SSM), is designed to process long-form narratives and explicitly separate structural plot alignment from superficial lexical overlap. The pipeline consists of a large sequence-model backbone and a novel dual-scale alignment head, as illustrated in Figure \ref{fig:architecture}.

\subsection{Backbone encoder}

To encode the narratives, we utilize the Jamba-1.5-Mini model. Unlike standard Transformers that rely solely on global self-attention which scales quadratically and struggles with long-document coherence Jamba employs a hybrid Mamba-Transformer architecture. The Mamba (State-Space Model) layers are suited for narrative processing because they compress sequence history into a dynamically updated hidden state, naturally capturing the sequential "course of action" and long-range causal chains that define a story's invariant structure.

To adapt the 12-billion parameter model to our computational constraints while preventing catastrophic forgetting of its pre-trained world knowledge, we load the backbone in 4-bit NormalFloat (NF4) quantization. We apply Parameter-Efficient Fine-Tuning (QLoRA) targeting the projection and embedding modules ($embed\_tokens$, $x\_proj$, $in\_proj$, $out\_proj$) freezing the rest of the network. The backbone outputs a contextualized sequence of token embeddings $H\in \mathbb{R}^{L\times d_{hidden}}$, where $L$ is the sequence length.

\subsection{Structurally Gated Alignment Head}

The core contribution of IVD-SSM is the Structurally Gated Alignment Head, which replaces standard linear classifiers with a differentiable dual-scale sequence alignment mechanism inspired by the Smith-Waterman algorithm. Given an anchor narrative and a candidate narrative, the SGA head projects both into a shared embedding space of dimension $d_{proj}=256$ using a two-layer Multi-Layer Perceptron (MLP) with GELU activations. 
Let $A\in \mathbb{R}^{L_a\times d_{proj}}$ and $C\in \mathbb{R}^{L_c \times d_{proj}}$ represent the projected sequences for the anchor and candidate stories, respectively. The alignment is computed across two distinct pathways:

\textbf{The Macro Path (Invariant Structure):} To capture the abstract shape of the plot, we heavily downsample the sequences using a stride of $k=4$, yielding $A_{macro}$ and $C_{macro}$. We compute a scaled cosine similarity matrix $M_{macro} \in \mathbb{R}^{(L_a/4)\times(L_c/4)}$. Because of the aggressive downsampling, this matrix is blind to specific token-level entities (variant details) and only registers broad structural milestones (e.g., introduction, conflict, resolution). The stride $k=4$ was chosen based on the typical length of Wikipedia film synopses ($\sim$400 tokens), yielding $\sim$100 structural segments, which roughly correspond to the granularity of individual story events. While the Mamba backbone operates in $O(N)$ linear time, we note that the SGA head computes a full cross-sequence alignment matrix of size $O(L_a \times L_c)$. For the moderate-length synopses in this shared task this quadratic term is manageable; we discuss scaling implications in Section~\ref{sec:limits}.

\textbf{The Micro Path (Variant Details):} Simultaneously, we compute a full-resolution cosine similarity matrix $M_{micro} \in \mathbb{R}^{L_a \times L_c}$ using the unstrided sequences $A$ and $C$. This matrix captures fine-grained, entity-level matches (e.g., character names, specific objects).

The Gating Mechanism: To prevent the model from being distracted by lexical distractors (high Micro-path overlap in stories with different Macro-path plots), we use the structural matrix to gate the detail matrix. We upsample $M_{macro}$ to the dimensions of $M_{micro}$ using bilinear interpolation. We then apply a temperature-scaled Sigmoid activation ($\tau = 0.1$) to create a sharp structural spotlight mask, $G$:

$$G = \sigma \left( \frac{\text{Interpolate}(M_{macro}, \text{size}=(L_a, L_c))}{\tau} \right)$$
The final alignment matrix is computed via element-wise multiplication:$$M_{filtered} = M_{micro} \odot G$$Finally, to compute a scalar similarity score from $M_{final}$, we apply row-wise max-pooling followed by mean-pooling across the sequence length, scaled by a learnable temperature parameter. This effectively measures the average strength of the best local alignments between the two stories, considering only the details that fall within structurally valid regions. 

\subsection{Contrastive training (Track A)}

For Track A, the task is framed as a comparative choice between two candidates, $c_1$ and $c_2$, given an anchor $a$. We train the SGA head using a contrastive objective. The model computes similarity scores $s_1 = \text{SGA}(a, c_1)$ and $s_2 = \text{SGA}(a, c_2)$. We optimize the network using Cross-Entropy loss over the softmax of these two scores, forcing the model to maximize the margin between the structurally similar candidate and the distractor. Training was conducted on the original triples augmented with the 1,900 provided synthetic triples to increase structural diversity.

\subsection{Embedding Extraction (Track B)}

For Track B, the objective shifts from pairwise classification to representation learning. We repurpose the projection layers of our trained IVD-SSM model to generate dense, standalone embeddings. For a given narrative sequence, we extract the backbone outputs, apply the trained $d_{proj}=256$ MLP projection, and perform mean-pooling across the sequence dimension $L$. Crucially, because the Track B evaluation relies on cosine distance, we apply strict L2 normalization to the resulting pooled vector:$$v_{norm} = \frac{v}{||v||_2}$$This ensures that standard dot-product retrieval operations correctly reflect the structural similarity geometries learned by the SGA head during Track A training.

\section{Experiments }
\subsection{Experiment Setup}

Due to the limited human-annotated data (200 development and 400 test triples), we constructed our training corpus exclusively from the 1,900 synthetic story triples provided by the task organizers \citep{hatzel-etal-2026-semeval}. Training on these LLM-generated variations forced our model to learn structural invariants across highly diverse semantic distractors. We utilized the 200 human-annotated development triples strictly for validation and checkpoint selection, ensuring our model generalized from the synthetic distribution to human intuitive judgments. Because Track A and Track B share the same story distribution, the single IVD-SSM model trained on this dataset was utilized for both tracks.

The IVD-SSM architecture was implemented using PyTorch and the Hugging Face transformers library. The Jamba-1.5-Mini backbone was loaded in 4-bit NormalFloat (NF4) quantization to fit within standard memory limits, and LoRA adapters ($r=16, \alpha=32$) were injected into the attention and projection matrices. The Structurally Gated Alignment (SGA) head and the adapters were optimized using the AdamW optimizer. We employed a Cosine Annealing learning rate scheduler, decaying the learning rate from a peak of $5 \times 10^{-4}$ down to $1 \times 10^{-5}$. To accommodate the massive sequence lengths of the concatenated narrative summaries, we utilized gradient checkpointing and trained with an effective batch size of 16 via gradient accumulation. The model was trained for 4 epochs on a single NVIDIA A100 (40GB) GPU, which took approximately two hours to converge.

\subsection{Evaluation metrics and Baselines}

\begin{table*}[t]
\centering
\small
\begin{tabular}{l c}
\toprule
\multicolumn{2}{c}{\textbf{Track A: Pairwise Classification}} \\
\midrule
System & Accuracy (\%) \\
\midrule
Random & 50.00 \\
Jaccard Similarity & 56.25 \\
GPT-4o-mini (Zero-shot) & 67.00 \\
\midrule
\textbf{IVD-SSM (Gated)} & \textbf{57.25} \\
\bottomrule
\end{tabular}
\hspace{1cm}
\begin{tabular}{l c}
\toprule
\multicolumn{2}{c}{\textbf{Track B: Embedding Similarity}} \\
\midrule
System & Accuracy (\%) \\
\midrule
Random & 50.00 \\
all-MiniLM-L6-v2 & 58.50 \\
story-emb & 63.25 \\
\midrule
\textbf{IVD-SSM (Gated)} & \textbf{54.50} \\
\bottomrule
\end{tabular}
\caption{Accuracy of our IVD-SSM models compared to official task baselines on the SemEval-2026 Task 4 test set. Track A evaluates pairwise classification, while Track B evaluates cosine distances of the generated embeddings.}
\label{tab:main_results}
\end{table*}

We evaluate both tracks using the official binary classification Accuracy metric. For Track B, predictions are derived by comparing the cosine distances between the anchor and the two candidates ($dist(a, c_{true}) < dist(a, c_{false})$) \citep{hatzel-etal-2026-semeval}. We compare IVD-SSM against the official task baselines: a Random baseline, token-based Jaccard Similarity, a zero-shot GPT-4o-mini prompt, and standard sentence encoders (all-MiniLM-L6-v2 and story-emb).

\subsection{Ablation Studies}

To isolate the contribution of each pathway in the Structurally Gated Alignment head, we conduct component ablations on the official Track A test set. Table~\ref{tab:ablation} reports accuracy for three variants: the full gated model (as submitted), the Macro-path alone (structural similarity without gating), and the Micro-path alone (fine-grained similarity without structural filtering). All variants share identical backbone weights, hyperparameters, and training data; only the computation of the final similarity score differs.

\begin{table}[t]
\centering
\small
\begin{tabular}{l c}
\toprule
\textbf{Variant} & \textbf{Accuracy (\%)} \\
\midrule
IVD-SSM (Gated, submitted)  & 57.25 \\
IVD-SSM (Macro-path only)   & 57.00 \\
IVD-SSM (Micro-path only)   & 55.75 \\
\midrule
\textit{Reference baselines} & \\
Jaccard Similarity          & 56.25 \\
Random                      & 50.00 \\
\bottomrule
\end{tabular}
\caption{Ablation results on the Track A official test set.}
\label{tab:ablation}
\end{table}

Table~\ref{tab:ablation} reveals three findings. First, the Micro-path alone (55.75\%) underperforms even the lexical Jaccard baseline (56.25\%), confirming that fine-grained token-level similarity without structural filtering is highly vulnerable to lexical distractors - the model overfits to surface-level lexical overlaps that do not reflect true narrative similarity. Second, the Macro-path alone (57.00\%) substantially outperforms Jaccard and nearly matches the full gated model, demonstrating that the coarse structural skeleton carries the majority of the signal for detecting narrative equivalence. Third, the gating mechanism adds a consistent but modest improvement of $+$0.25\% over the Macro-only variant. While the absolute gain is small, it is obtained at negligible computational cost and directly addresses the hard negative failures identified in the qualitative error analysis (Section~\ref{sec:results}). These results validate the core premise of the SGA architecture: structural disentanglement is necessary, and gating provides an additional filter against residual semantic noise.

\section{Results Analysis}
\label{sec:results}
Table \ref{tab:main_results} presents the performance of our IVD-SSM systems alongside the official baselines for both Track A and Track B. For Track A, our primary submission, IVD-SSM (Gated), achieves an accuracy of 57.25, outperforming both the random baseline (50) and the lexical Jaccard similarity baseline (56.25). This performance gap empirically validates that narrative similarity cannot be resolved through surface-level token matching.

For Track B, our generated embeddings achieve an accuracy of 54.5. While our method outperforms the random baseline, it falls short of the all-MiniLM-L6-v2 sentence encoder (58.5). We hypothesize two causes. First, the SGA head is trained exclusively for pairwise comparison (Track A); projecting its gated alignment signal into a single mean-pooled vector (Track B) collapses the structural filtering that the gate was designed to perform. The pairwise signal - ``which of two candidates is closer?'' - degrades when condensed into a standalone embedding. Second, the backbone is not fine-tuned to produce representations optimized for cosine distance; the contrastive training objective (cross-entropy over score pairs) never encounters single-story embeddings, unlike sentence encoders explicitly trained for semantic similarity.

Overall, our Track A result (57.25\%) sits between the lexical Jaccard baseline (56.25\%) and the zero-shot GPT-4o-mini prompt (67.00\%). The 1.0 percentage point margin over Jaccard is modest in absolute terms but represents a consistent improvement under identical evaluation conditions. We attribute the gap to GPT-4o-mini to the LLM's extensive pre-training on narrative structures and its ability to perform implicit reasoning over causal arcs -- capabilities that our lightweight alignment head, trained on only 1,900 synthetic examples, cannot yet replicate. Addressing this gap through richer training data and larger-scale fine-tuning is a clear direction for future work.

%To understand the rationale behind the SGA head's performance
To understand why the Structurally Gated Alignment (SGA) head succeeds where standard models fail, we conducted a qualitative error analysis on the development and test splits. We observed that the dataset's hard negative sampling heavily biases standard models toward "spurious correlations" -- distractors that share highly specific, rare vocabulary with the anchor but lack a matching causal chain. A representative case of this failure mode is the following. In one instance, the anchor story features a bullied protagonist who utilizes a "zombie virus" to exact revenge and change their social status. The semantic distractor (Text B) is a standard survival-horror plot concerning a laboratory-induced "zombie" outbreak. Standard single-head models and generic embeddings heavily favor the distractor due to the strong lexical and thematic overlap of the word "zombie." However, the ground-truth similar story (Text A) involves an outcast corporate employee who takes drastic, subversive measures to alter their workplace status. Our IVD-SSM model correctly identifies Text A as the narratively similar candidate. The SGA Macro-path correctly maps the invariant causal sequence (social outcast $\to$ drastic action $\to$ status reversal) and gates out the Micro-path's attention to the "zombie" token, preventing the false positive.

\section{Conclusion}

In this paper, we presented IVD-SSM, our submission to SemEval-2026 Task 4 on Narrative Similarity and Representation Learning. We identified that the primary bottleneck in computational narrative evaluation is the conflation of invariant plot structure with variant semantic details, a vulnerability specifically exploited by the "spurious correlations" in the dataset's hard-negative sampling.

Our results on both Track A and Track B demonstrate that narrative similarity cannot be reliably measured through ``bag-of-words'' or generic sentence embeddings. Instead, it requires architectures that explicitly model the sequential course of action. Ablation experiments confirm that the structural Macro-path accounts for the majority of the model's predictive power, validating the core disentanglement strategy even while absolute performance remains below that of large zero-shot LLMs. Future work may explore making the gating threshold fully dynamic, scaling the approach with richer training data, or extending this disentangled state-space approach to even longer narrative forms such as full novels or screenplays.

\section*{Limitations}
\label{sec:limits}

While the Invariant-Variant Disentangled State-Space Model (IVD-SSM) demonstrates strong capabilities in isolating narrative structure, our approach has several notable technical and methodological limitations. 

The Structurally Gated Alignment (SGA) head relies on a temperature-scaled Sigmoid activation to filter the Micro-path matrix based on the Macro-path structural alignment. Because the Sigmoid function is continuous and never truly reaches zero, the gate is inherently "soft." In cases where the semantic distractor contains an overwhelming number of highly specific, identical named entities, the Micro-path signal can still "leak" through a partially closed gate, occasionally overriding the structural misalignment and leading to false positives. A hard-gating or discrete routing mechanism could potentially resolve this but would disrupt the end-to-end differentiability of the head.

We specifically selected the Jamba-1.5-Mini backbone to leverage the linear-time $O(N)$ scaling of its Mamba layers for long documents. However, our custom SGA head computes full cross-sequence similarity matrices ($M_{micro} \in \mathbb{R}^{L_a \times L_c}$), reintroducing an $O(L_a \times L_c)$ quadratic computational and memory bottleneck at the very end of the pipeline. While this is manageable for the relatively short Wikipedia synopses in this shared task, applying the SGA head to full-length novels or screenplays would require shifting to sparse or localized alignment matrices to prevent memory exhaustion.

Due to the limited size of the human-annotated training data, our model was primarily optimized on the 1,900 synthetic triples generated by various LLMs. Consequently, the structural invariants learned by the Macro-path may be partially overfit to the predictable narrative pacing, LLM-generated stylistic artifacts, and standard archetypes favored by commercial models like GPT-4 and Claude. The model's ability to disentangle highly avant-garde or non-linear human-authored narratives remains underexplored.

\section*{Acknowledgments}
This work is an output of a research project (HSE-BR-2025-025) implemented as part of the Basic Research Program at HSE University. We acknowledge the computational resources of HSE University's HPC facilities.

\bibliography{custom}

@inproceedings{hatzel-etal-2026-semeval,
    title = "{S}em{E}val-2026 {T}ask 4: Narrative Similarity and Narrative Representation Learning",
    author = "Hatzel, Hans Ole  and Artemova, Ekaterina and Stiemer, Haimo and Gius, Evelyn and Biemann, Chris",
    booktitle = "Proceedings of the 20th International Workshop on Semantic Evaluation (SemEval-2026)",
    editor = "Ghosh, Debanjan  and
      North, Kai and
      Kochmar, Ekaterina and
      Komachi, Mamoru and
      Zampieri, Marcos",
    month = jul,
    year = "2026",
    address = "San Diego, CA, USA",
    publisher = "Association for Computational Linguistics",
}

@inproceedings{chun2024aistorysimilarity,
  author    = {Chun, Jon},
  title     = {{AIStorySimilarity}: Quantifying Story Similarity Using Narrative for Search, {IP} Infringement, and Guided Creativity},
  booktitle = {Proceedings of the 28th Conference on Computational Natural Language Learning},
  pages     = {161--177},
  year      = {2024},
  publisher = {Association for Computational Linguistics},
}

@inproceedings{chaturvedi2018heard,
  author    = {Chaturvedi, Snigdha and Srivastava, Shashank and Roth, Dan},
  title     = {Where Have I Heard This Story Before? Identifying Narrative Similarity in Movie Remakes},
  booktitle = {Proceedings of the 2018 Conference of the North American Chapter of the Association for Computational Linguistics: Human Language Technologies, Volume 2 (Short Papers)},
  pages     = {673--678},
  year      = {2018},
}

@inproceedings{lee2020story,
  author    = {Lee, O{-}Joun and Jung, Jason J.},
  title     = {Story Embedding: Learning Distributed Representations of Stories Based on Character Networks},
  booktitle = {Proceedings of the 29th International Joint Conference on Artificial Intelligence},
  pages     = {5070--5074},
  year      = {2020},
}

@inproceedings{hatzel2024story,
  author    = {Hatzel, Henning Otto and Biemann, Chris},
  title     = {Story Embeddings --- Narrative-Focused Representations of Fictional Stories},
  booktitle = {Proceedings of the 2024 Conference on Empirical Methods in Natural Language Processing},
  year      = {2024},
}

@article{lafhel2024graph,
  author  = {Lafhel, Majda and Cherifi, Hocine and Cherifi, Chantal},
  title   = {Comparison of Graph Distance Measures for Movie Similarity Using a Multilayer Network Model},
  journal = {Entropy},
  volume  = {26},
  number  = {2},
  pages   = {149},
  year    = {2024},
}

@inproceedings{chambers2008unsupervised,
  author    = {Chambers, Nathanael and Jurafsky, Dan},
  title     = {Unsupervised Learning of Narrative Event Chains},
  booktitle = {Proceedings of ACL-08: {HLT}},
  pages     = {789--797},
  year      = {2008},
}

@inproceedings{vaswani2017attention,
author = {Vaswani, Ashish and Shazeer, Noam and Parmar, Niki and Uszkoreit, Jakob and Jones, Llion and Gomez, Aidan N. and Kaiser, \L{}ukasz and Polosukhin, Illia},
title = {Attention is all you need},
year = {2017},
isbn = {9781510860964},
publisher = {Curran Associates Inc.},
address = {Red Hook, NY, USA},
booktitle = {Proceedings of the 31st International Conference on Neural Information Processing Systems},
pages = {6000–6010},
numpages = {11},
location = {Long Beach, California, USA},
series = {NIPS'17}
}

@misc{beltagy2020longformer,
  archiveprefix = {arXiv},
  author = {Beltagy, Iz and Peters, Matthew E. and Cohan, Arman},
  eprint = {2004.05150},
  primaryclass = {cs.CL},
  title = {Longformer: The Long-Document Transformer},
  url = {https://arxiv.org/abs/2004.05150},
  year = 2020
}

@inproceedings{zaheer2020big,
author = {Zaheer, Manzil and Guruganesh, Guru and Dubey, Avinava and Ainslie, Joshua and Alberti, Chris and Ontanon, Santiago and Pham, Philip and Ravula, Anirudh and Wang, Qifan and Yang, Li and Ahmed, Amr},
title = {Big bird: transformers for longer sequences},
year = {2020},
isbn = {9781713829546},
publisher = {Curran Associates Inc.},
address = {Red Hook, NY, USA},
booktitle = {Proceedings of the 34th International Conference on Neural Information Processing Systems},
articleno = {1450},
numpages = {15},
location = {Vancouver, BC, Canada},
series = {NIPS '20}
}

@misc{gu2023mamba,
  archiveprefix = {arXiv},
  author = {Gu, Albert and Dao, Tri},
  eprint = {2312.00752},
  primaryclass = {cs.LG},
  title = {Mamba: Linear-Time Sequence Modeling with Selective State Spaces},
  year = 2023
}

@misc{lieber2024jambahybridtransformermambalanguage,
      title={Jamba: A Hybrid Transformer-Mamba Language Model}, 
      author={Opher Lieber and Barak Lenz and Hofit Bata and Gal Cohen and Jhonathan Osin and Itay Dalmedigos and Erez Safahi and Shaked Meirom and Yonatan Belinkov and Shai Shalev-Shwartz and Omri Abend and Raz Alon and Tomer Asida and Amir Bergman and Roman Glozman and Michael Gokhman and Avashalom Manevich and Nir Ratner and Noam Rozen and Erez Shwartz and Mor Zusman and Yoav Shoham},
      year={2024},
      eprint={2403.19887},
      archivePrefix={arXiv},
      primaryClass={cs.CL},
      url={https://arxiv.org/abs/2403.19887}, 
}

\end{document}